\documentclass[sigconf]{acmart}
\usepackage{booktabs}
\usepackage{graphicx}
\usepackage[normalem]{ulem}
\useunder{\uline}{\ul}{}
\usepackage{longtable}
\usepackage{stfloats}
\usepackage{amsfonts}
\usepackage{booktabs}
\usepackage{array}
\usepackage{longtable}
\usepackage{subfigure}
\usepackage{amsmath}
\usepackage{color, colortbl}
\usepackage{bm}
\usepackage{appendix}
\usepackage{diagbox}
\usepackage{multirow}
\usepackage{lscape}
\usepackage[shortlabels]{enumitem}
\definecolor{Gray}{gray}{0.9}

\newcommand{\tabincell}[2]{\begin{tabular}{@{}#1@{}}#2\end{tabular}}

\AtBeginDocument{%
  \providecommand\BibTeX{{%
    \normalfont B\kern-0.5em{\scshape i\kern-0.25em b}\kern-0.8em\TeX}}}
\newcommand{\name}{\textsc{DISCOS}}

\setcopyright{iw3c2w3}
\copyrightyear{2021}
\acmYear{2021}
\acmConference[WWW '21]{Proceedings of the Web Conference 2021}{April 19--23, 2021}{Ljubljana, Slovenia}
\acmBooktitle{Proceedings of the Web Conference 2021 (WWW '21), April 19--23, 2021, Ljubljana, Slovenia}
\acmPrice{}
\acmDOI{10.1145/3442381.3450117}
\acmISBN{978-1-4503-8312-7/21/04}

\begin{document}

\title{\name: Bridging the Gap between Discourse Knowledge and Commonsense Knowledge}

\author{Tianqing Fang}
\email{tfangaa@cse.ust.hk}
\affiliation{%
  \institution{CSE, HKUST}
  \city{Hong Kong}
  \state{China}
}
\author{Hongming Zhang}
\email{hzhangal@cse.ust.hk}
\affiliation{%
  \institution{CSE, HKUST}
  \city{Hong Kong}
  \state{China}
}
\author{Weiqi Wang}
\email{wwangbw@connect.ust.hk}
\affiliation{%
  \institution{CSE, HKUST}
  \city{Hong Kong}
  \state{China}
}
\author{Yangqiu Song}
\email{yqsong@cse.ust.hk}
\affiliation{%
  \institution{CSE, HKUST}
  \city{Hong Kong}
  \state{China}
}
\author{Bin He}
\email{hebin.nlp@huawei.com}
\affiliation{%
  \institution{Huawei Noah’s Ark Lab}
  \state{China}
}



\begin{abstract}

Commonsense knowledge is crucial for artificial intelligence systems to understand natural language.
Previous commonsense knowledge acquisition approaches typically rely on human annotations (for example, ATOMIC) or text generation models (for example, COMET.) Human annotation could provide high-quality commonsense knowledge, yet its high cost often results in relatively small scale and low coverage. On the other hand, generation models have the potential to automatically generate more knowledge. Nonetheless,  machine learning models often fit the training data well and thus struggle to generate high-quality novel knowledge.
To address the limitations of previous approaches, in this paper, we propose an alternative commonsense knowledge acquisition framework \name{} (from DIScourse to COmmonSense), which automatically
populates 
expensive complex commonsense knowledge to more affordable linguistic knowledge resources. 
Experiments demonstrate that we can successfully convert discourse knowledge about eventualities from ASER, a large-scale discourse knowledge graph, into 
\textit{if-then} commonsense knowledge defined in ATOMIC without any additional annotation effort. 
Further study suggests that DISCOS significantly outperforms previous supervised approaches in terms of novelty and diversity with comparable quality. 
In total, we can acquire 3.4M ATOMIC-like inferential commonsense knowledge by populating ATOMIC on the core part of ASER. Codes and data are available at \url{https://github.com/HKUST-KnowComp/DISCOS-commonsense}.

\end{abstract}

\maketitle

\section{Introduction}



Understanding commonsense knowledge has long been one of the ultimate goals of the artificial intelligence field.
To achieve that goal, many efforts have been devoted to acquire commonsense knowledge.
For example, ConceptNet~\cite{speer2017conceptnet} (originally known as Open Mind Common Sense (OMCS)~\cite{liu2004conceptnet}) and OpenCyc \cite{lenat1989building} leverage expert annotation and integration of existing knowledge bases to acquire high-quality commonsense knowledge about human-defined relations.
The majority of these relations are factoid commonsense such as {\it isA}, {\it partOf}, {\it attributeOf}, etc.
Recently, the focus of sequences of events and the social commonsense relating to them has drawn a lot of attention.
ATOMIC~\cite{bosselut-etal-2019-comet} is such a knowledge base about inferential knowledge organized as typed \textit{if-then} relations with variables being events and states.
Different from traditional knowledge bases, events and states are usually more loosely-structured texts to handle diverse queries of commonsense represented by our natural language.
Though being potentially useful for solving commonsense reasoning applications, such kind of commonsense knowledge also brings new challenges for machines to acquire new knowledge of the similar type and make inferences. 

First, the knowledge acquired by ATOMIC are based on crowdsourcing, which are relatively more expensive than other automatic information extraction methods. 
To overcome this problem, COMET~\cite{bosselut-etal-2019-comet} is proposed to finetune a large pre-trained language model (i.e., GPT~\cite{radford2018improving}) with existing commonsense knowledge bases (for example, ATOMIC) such that they can automatically generate reasonable commonsense knowledge.
Even though COMET can generate high-quality complex commonsense knowledge with the supervised approach, it tends to 
fit the training data too well to generate novel concepts.
This is usually referred to as a \textit{selection bias} problem in statistical analysis~\cite{NBERw0172,Zadrozny04}.

\begin{figure*}[t]
    \centering
        \includegraphics[width=0.7\textwidth]{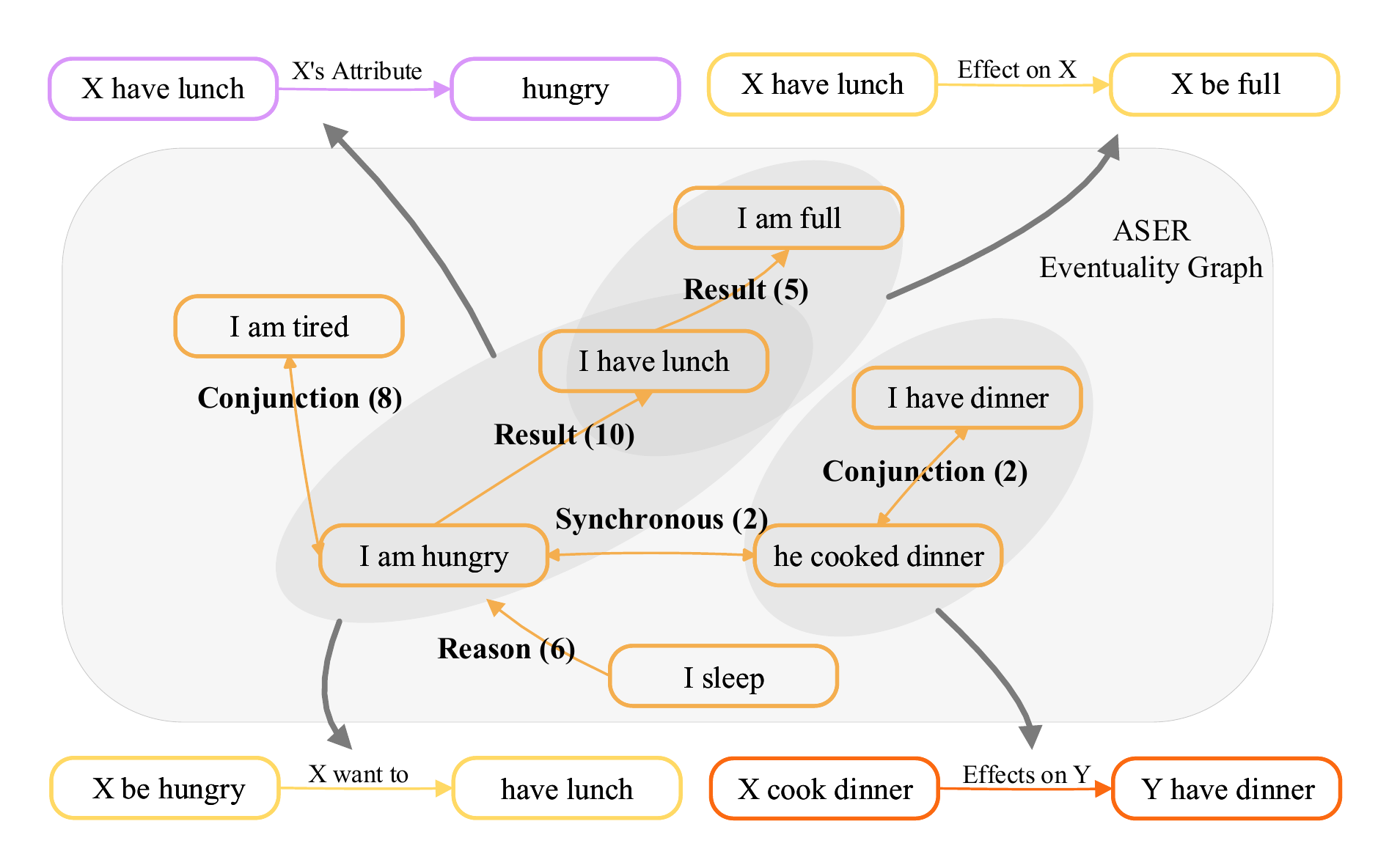}
        \vspace{-2ex}
    \caption{An illustration of \name{}. Eventualities from ASER are connected by directed edges denoting the corresponding discourse relationships. DISCOS aims to transform the discourse edges in ASER to \textit{if-then} commonsense edges. For example, an ASER edge (``I am hungry,'' \textit{Result}, ``I have lunch'') will be transformed to (\textit{if} ``X be hungry,'' \textit{then} \textit{X Want to},  ``have lunch'') commonsense tuple. Other discourse edges can also entail other commonsense relations. } \label{fig:ASER-sketch}
\end{figure*}

On the other hand, although information extraction may be also subject to {\it reporting bias}~\cite{GordonD13}, where the frequencies may not truly reflect the relative likelihood in the real-world, it can provide a lot of candidate examples that can be evaluated by a machine learning model trained on human annotated data.
For example, ASER~\cite{zhang2020aser} uses frequent syntactical patterns to extract eventualities (such as activities or processes, events, and states) in a dependency parse of a sentence. 
Then it forms linguistic relations between eventualities based on discourse markers (such as ``and,'' ``but,'' etc.)
Such an automatic extraction approach can easily scale up to two orders of magnitude larger than human annotations.
However, it is not trivial to leverage such a knowledge resource. 
First, ASER and ATOMIC have different formats.
As shown in Figure~\ref{fig:ASER-sketch}, the knowledge in ASER are mostly natural language, for example, ``I am hungry,'' whereas in ATOMIC, person entities are mostly aggregated, for example, ``Person X be hungry.''
Thus, aligning ATOMIC with ASER needs additional efforts on deeply exploring both knowledge bases. 
Second, while some discourse relations extracted in ASER can naturally reflect the {\it if-then} relations, they are not all valid for each of the \textit{if-then} relations with variables being events and states. 
For example, a {\it Succession} relation in ASER, which is usually extracted by connectives such as ``after'' and ``once,'' cannot be used as a candidate relation for the {\it Stative} relation in ATOMIC, because, by definition, the \textit{Stative} represents the state of the agent at the same time or before the base event happens, which is opposite from the chronological order of {\it Succession}.

Last but not least, although it is widely accepted that the graph substrucure can be useful for making predictions and inferences in entity-centric knowledge graphs~\cite{Teru2020SubgraphKG}, existing commonsense knowledge based models~\cite{bosselut-etal-2019-comet,petroni-etal-2019-language} still treat the prediction as a translational problem for the triplets in the knowledge base and do not consider the subgraph structures. 
It is also not trivial to leverage graph structures in commonsense knowledge acquisition.
First, there is no existing graph structure in ATOMIC, as the labeling procedure only considers the head, the tail, and their relations.
There are few overlaps between heads and tails given both can be arbitrary texts.
The heads and tails can form a bipartite graph but graph convolution in such a graph may not provide additional information compared to direct representation learning for nodes, because tails can be conditionally independent given a head.
However, with ASER, which is a more structural knowledge graph, it is possible to perform more complicated reasoning over the substructures.
Second, as we mentioned that heads and tails are loosely-structured texts in both ATOMIC and ASER, a contextualized representation model should be applied to them for better representations.
As a result, 
when developing a graph-based model for commonsense acquisition, both the scalability and effectiveness should be carefully considered.

To address the above challenges, in this paper, we propose a new commonsense knowledge acquisition framework, \name{} (from DIScourse knowledge to COmmonSense knowledge), which leverages the large-scale eventuality-centric discourse knowledge in ASER to enrich the inferential commonsense knowledge in ATOMIC.
Figure~\ref{fig:ASER-sketch} shows an example of the results.
Different from existing mechanisms such as tail node prediction adopted in COMET \cite{bosselut-etal-2019-comet} and link prediction in knowledge base completion tasks used by KG-Bert~\cite{yao2019kg}, we propose a knowledge base population approach for \name{}.
This can be done by first mapping ATOMIC nodes to ASER nodes, and then performing a transductive learning algorithm which is based on both contextualized text representation (i.e., BERT~\cite{devlin-etal-2019-bert}) and a graph-related representation (i.e., graph neural networks  \cite{hamilton2017inductive}) to aggregate neighborhood information to jointly make decisions on whether we can populate the ATOMIC relations to a pair of ASER nodes.
Experiments demonstrate that the proposed model inherits the advantage of both text and graph representation learning models.
Compared with the learning method trained on ATOMIC only, we significantly improve the novelty and diversity of the acquired commonsense knowledge, with comparable accuracy.
Extensive analysis are conducted to analyze the strengths and limitations of \name{}. 

Our contributions can be summarized as follows.

$\bullet$ We formulate commonsense acquisition as a Commonsense Knowledge Graph Population (CKGP) task, and propose a novel framework, \name{}, to populate the inferential {\it if-then} commonsense knowledge in ATOMIC to an eventuality-centric discourse knowledge graph ASER.

$\bullet$ In DISCOS, we develop a model named \textsc{BertSAGE} to jointly leverage the textual representation and graph representation to discriminate commonsense knowledge. This model can be used as a general approach for commonsense knowledge graph population.

$\bullet$ We not only systematically evaluate our framework with commonly used evaluation metrics such as novelty and accuracy using both benchmark dataset and human evaluations, but also thoroughly analyze our models and results as well as the patterns shown in both ATOMIC and ASER to demonstrate that incorporating information extraction results in ASER to enrich the {\it if-then} relations can indeed provide larger-scale qualified commonsense knowledge.

We organize the rest of this paper as follows. Section \ref{section:related-work} provides a systematic review on commonsense knowledge bases and commonsense knowledge reasoning. Section \ref{section:preliminary} introduces the definition of our commonsense acquisition task, as well as basic knowledge about discourse knowledge graph and the source commonsense knowledge base that we use, which are ASER and ATOMIC, respectively. The details of \name{} are presented in Section \ref{section:methods}. Section \ref{section:experiments} presents the experimental settings and results of the experiments. The corresponding ablation studies are illustrated in Section \ref{section:ablation}. Some case studies and discussions about the effects of ASER are presented in Section \ref{section:case-discussion}. Finally, the paper is concluded in Section \ref{section:conclusion}.

\section{Related Work}\label{section:related-work}

Commonsense knowledge spans a large range of human experience including spatial, physical, social, temporal, and psychological aspects of everyday life \cite{liu2004conceptnet}. 
Commonsense knowledge has been shown to be crucial for many natural language understanding tasks including question answering \cite{wang2017explicit, talmor2019commonsenseqa}, dialogue understanding and generation \cite{young2018augmenting, wu2020diverse, wang2018yuanfudao}, event prediction \cite{ding2019event, tandon2018reasoning}, and story understanding and generation \cite{guan2019story, yang2019knowledgeable, guan2020knowledge}.

To bridge the gap between world knowledge and natural language understanding (NLU) systems, several large-scale CommonSense Knowledge Bases (CSKB) are proposed  \cite{liu2004conceptnet,speer2017conceptnet,sap2019atomic}. 
The Open Mind Common Sense project (OMCS)~\cite{liu2004conceptnet} defined 20 commonsense relations (for example, \textit{UsedFor}, \textit{AtLocation}) and manually annotate over 600K assertions.
On top of that, ConceptNet 5.0~\cite{speer2017conceptnet} extended it to 36 relations over 8M structured nodes and 21M edges by incorporating more knowledge from other resources like WordNet \cite{miller1998wordnet} and OpenCyc \cite{lenat1989building}.
However, even with these great efforts, ConceptNet still cannot cover all commonsense knowledge. For example, ConceptNet is limited to entity-centric nodes and does not provide much commonsense knowledge about daily events. 
To fill this gap, ATOMIC \cite{sap2019atomic} was proposed to gather the everyday commonsense knowledge about events.
Specifically, ATOMIC defined 9 relations and crowdsourced 880K assertions. The nodes in ATOMIC are eventuality-centric, i.e., they are verb phrases or a complete sentence, which is typically more complicated than ConceptNet and provides richer information about events. In addition, the extended work of ATOMIC and COMET, ATOMIC-2020~\cite{hwang2020comet} reformulated the commonsense relations in ATOMIC into three parts, i.e., social interaction, physical-entity relations, and event-centered  relations. 
New annotations are provided for novel relations.
While ATOMIC mostly presents non-contextual causal commonsense for daily events, GLUCOSE~\cite{mostafazadeh-etal-2020-glucose} formalized ten causal dimensions over stories that contains fruitful context. The ten causal dimensions are not only person-centric, but can also involve non-person entities like places and objects.

One common limitation of these knowledge graphs is that the human annotation can be expensive and thus it is infeasible to further scale them up to cover all commonsense knowledge.
To address the limitation of human annotation, many recent works tried to acquire commonsense knowledge automatically.
For example, several recent works have been focusing on mining commonsense knowledge using pre-trained language models \cite{davison-etal-2019-commonsense, Bhagavatula2020Abductive}.
LAMA \cite{petroni-etal-2019-language} manually created cloze statements from CSKB, and predicted the clozes using the BERT model. They found that the pre-trained BERT itself contains much commonsense knowledge without fine-tuning.
COMET \cite{bosselut-etal-2019-comet} used unidirectional language model GPT \cite{radford2018improving} to generate new commonsense knowledge for ConceptNet and ATOMIC, showing good performance based on human evaluation. However, the ability to generate novel and diverse commonsense knowledge is limited due to the nature of encoder-decoder framework and beam search, as reported from experiments \cite{bosselut-etal-2019-comet, cohen2019empirical}. 
Furthermore, a guided generation model with lexically-constrained decoding for cause and effects is designed to explore the causal relations in the Common Crawl corpus~\cite{buck-etal-2014-n}. Combining the constructed cause-effect graph and a neural generative model, the framework can perform better reasoning than pre-trained language models.

Besides generation methods, TransOMCS \cite{zhang2020transomcs} first formalized the task of mining commonsense knowledge from linguistic graphs. They automatically extracted patterns from ASER \cite{zhang2020aser}, a large linguistic graph constructed from various web corpora, using ConceptNet as seed commonsense knowledge, and retrieved high-quality commonsense tuples based on a ranking model. However, due to the limitation of pattern mining, TransOMCS can only deal with short, canonical phrases like the nodes in ConceptNet, and cannot be generalized to free-text and complicated linguistic patterns.


\begin{figure}[t]
    \centering
        \includegraphics[width=8.5cm]{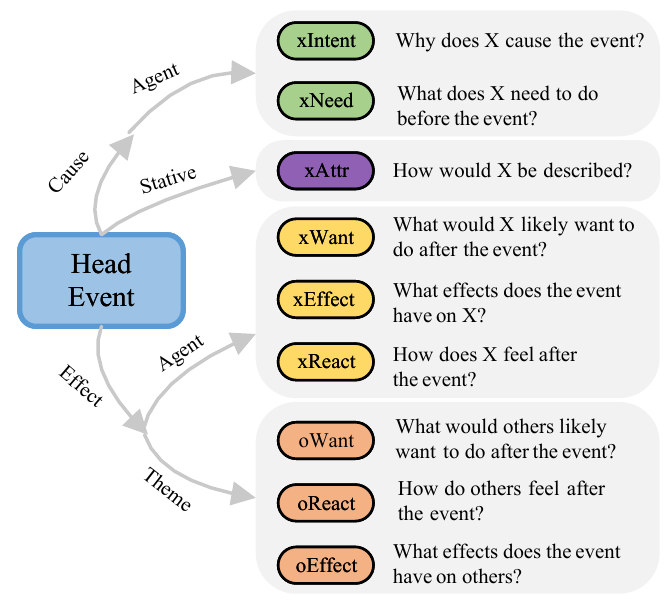}
    \caption{ATOMIC relation definition. The relations are categorized based on chronological order and the subject of events. (1) \textit{cause\_agent}: What causes the agent (X) to do the events. (2) \textit{stative}: What is the state of the agent (X). (3) \textit{effect\_agent}: What are the effects on the agent (X). (4) \textit{effect\_theme}:  What are the effects on the theme (Others).} \label{fig:ATOMIC-relation}
\end{figure}

Besides automatically generating commonsense tuples, 
another line of work is treating commonsense acquisition as a knowledge base completion task~\cite{li2016commonsense}.
Throughout the years, many techniques including LSTM and aggregation~\cite{li2016commonsense,saito2018commonsense}, pre-trained CNN \cite{malaviya2020exploiting}, and inductive learning \cite{wang2020inductive} have been proposed to model the commonsense relations between objects.
Even though these models cannot be directly applied to generate novel objects, they can serve as good classification models to tell whether a new generated commonsense assertion is plausible or not.
In this paper, to leverage the advantages of both the generation and classification models, we map ATOMIC with ASER to generate a large scale knowledge graph where the relations acquired by ATOMIC will be used as a supervision signal to train a graph representation learning model. Then the model is used to predict edges introduced in ASER to acquire more similar relations as ATOMIC. Our approach can be essentially regarded as a knowledge base population task~\cite{JiG11}.
Compared to the knowledge base completion task, which assumes the nodes in a knowledge base are fixed and only predict new edges,  a knowledge base population task can introduce both new nodes and new edges to an existing knowledge base.


\begin{figure*}[t]
    \centering
        \includegraphics[width=17cm]{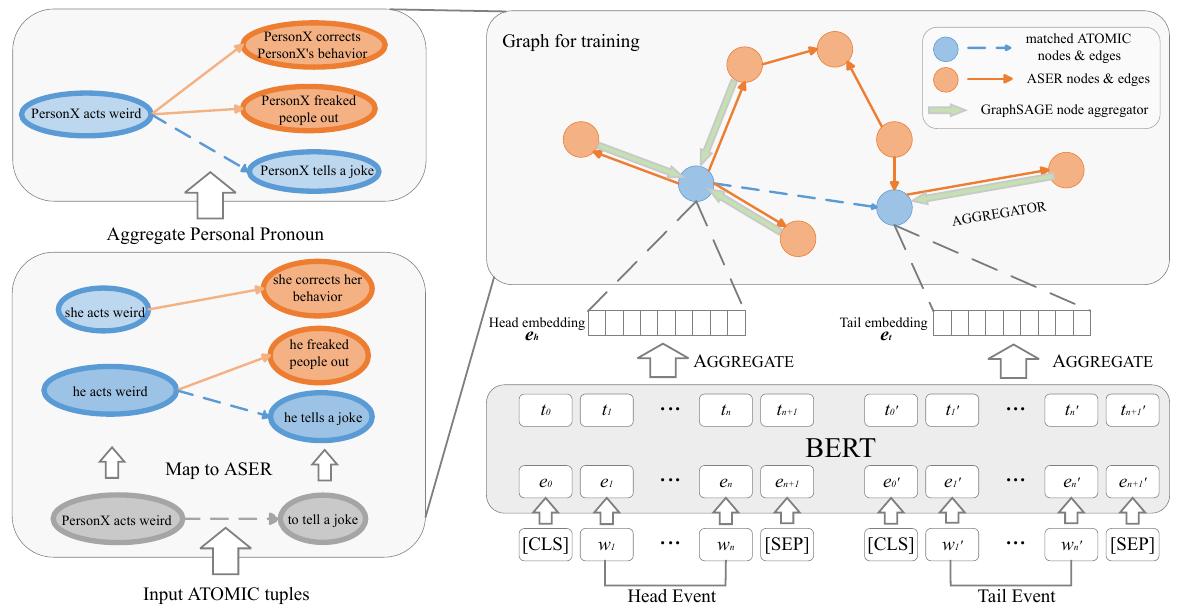}
    \caption{DISCOS overview. 
    First, ATOMIC tuples are mapped to ASER format to acquire candidate commonsense knowledge neighbors from discourse edges in ASER. Then \textsc{BertSAGE} is used to discriminate whether a $(h, r, t)$ tuple is plausible or not.
    } \label{fig:model-overview}
\end{figure*}

\section{Preliminaries}\label{section:preliminary}
In this section, we introduce basic formulations of the commonsense knowledge acquisition task and the resource that are used. We use ATOMIC~\cite{sap2019atomic} as the seed commonsense knowledge graph because it contains rich complex commonsense knowledge, and ASER~\cite{zhang2020aser} as the discourse knowledge graph.
Details are as follows.


\subsection{Task Definition}\label{sec:task-definition}
The task of acquiring commonsense knowledge from discourse knowledge graphs is defined as a \textbf{Commonsense Knowledge Graph Population (CKGP)} task.
The seed commonsense knowledge graph is denoted as $\mathcal{C}=\{(h, r, t) | h \in \mathcal{H}, r\in \mathcal{R}, t\in \mathcal{T}\}$, where $\mathcal{H}$, $\mathcal{R}$, and $\mathcal{T}$ are the set of the heads, relations, and tails, respectively. Suppose we have another much larger knowledge graph extracted from texts via discourse relations, denoted as $\mathcal{G}=(\mathcal{V}, \mathcal{E})$, where $\mathcal{V}$ is the set of all vertices and $\mathcal{E}$ is the set of edges, storing the discourse relations among eventualities.

The CKGP model is trained using a link prediction task 
over the aligned graph $\mathcal{G}^c$ that contains both the edges from $\mathcal{G}$ and $\mathcal{C}$.
The ground truth edges are the corresponding edges from the source commonsense knowledge graph $\mathcal{C}$. 
After learning the edge information from $\mathcal{C}$, in the inference process, the model is asked to predict plausible tail $t$ given head $h$ and relation $r$ as input.
Specifically, there are two settings for the inference process: (1) \textit{Existing head}: Predict tails given head $h$ from the original $\mathcal{C}$, (2) \textit{Novel head}:  Predict tails given head $h$ from $\mathcal{G}$ that does not appear in $\mathcal{C}$. While previous works~\cite{sap2019atomic, bosselut-etal-2019-comet} adopt the first setting, we argue that the second setting can generate commonsense knowledge in a much larger scale, considering that $\mathcal{G}$ is much larger.



\subsection{ATOMIC}

We adopt ATOMIC \cite{sap2019atomic} as the seed commonsense knowledge $\mathcal{C}$. 
ATOMIC consists of 880K tuples across nine relations about day-to-day \textit{if-then} commonsense knowledge (for example, \textit{if} X feels hungry, \textit{then} X wants to have lunch.)
Different from structured or canonical knowledge bases, the nodes in ATOMIC are in the form of free-text, which is more expressive in representing everyday commonsense but also makes the matching and generation harder. 
As shown in Figure \ref{fig:ATOMIC-relation}, the nine relation types span over four categories, which are classified based on the order of time and subject of the events.
Detailed illustrations can be found in Figure \ref{fig:ATOMIC-relation}.


\subsection{ASER}

ASER \cite{zhang2020aser}, a large-scale eventuality-centric knowledge graph that provides explicit discourse relationships between eventualities, is used as the source of discourse knowledge graph $\mathcal{G}$. 
We use the core part of ASER, which consists of 15 discourse relation types, and 10M discourse edges among 27M eventualities. As illustrated in Figure \ref{fig:ASER-sketch}, the discourse relation (``I am hungry,'' \textit{Result}, ``I have lunch'') can  be potentially transformed to \textit{if-then} commonsense knowledge, i.e., (``X be hungry,'' \textit{X want to}, ``have lunch.'')

\section{DISCOS}\label{section:methods}

The overall framework of DISCOS is shown in Figure \ref{fig:model-overview}. First, the subjects of events in ATOMIC and ASER are quite different, where in ATOMIC the subjects are placeholders like ``\textit{PersonX}'' and ``\textit{PersonY},'' while in ASER they are concrete personal pronouns like ``she'' and ``he.'' So, in order to align the two resources to perform Commonsense Knowledge Graph Population, we first map all heads and tails in $\mathcal{C}$ (ATOMIC) into $\mathcal{G}$ (ASER). Formally, we need a mapping function $M(s)$ to map the input string $s$ into the same format of nodes in $\mathcal{G}$, 
such that we can find as many $(h, r, t) \in \mathcal{C}$ tuples as possible that can be matched to $\mathcal{G}$ using $M(h)$ and $M(t)$ operations. 
Next, we leverage a rule $D(v, r), v\in \mathcal{V}, r \in \mathcal{R}$, to select candidate discourse edges in $\mathcal{G}$, given a node $v=M(h), h\in \mathcal{H}$ and a commonsense relation $r$. After finding all candidate discourse edges under relation $r$, denoted as $\mathcal{L}(r)=\{(u, v) | (u, v)\in \mathcal{E}\}$, we employ a novel commonsense knowledge population model, \textsc{BertSAGE}, to score the plausibility of the candidate commonsense tuple $(v, r, u)$.
This framework is not restricted to the resource of ATOMIC and ASER, but can be well generalized to other resources, as one can change the mapping rules accordingly and use the \textsc{BertSAGE} model flexibly.
Details about each step are introduced as follows.

\begin{figure}[!t]
    \centering
        \includegraphics[width=0.48\textwidth]{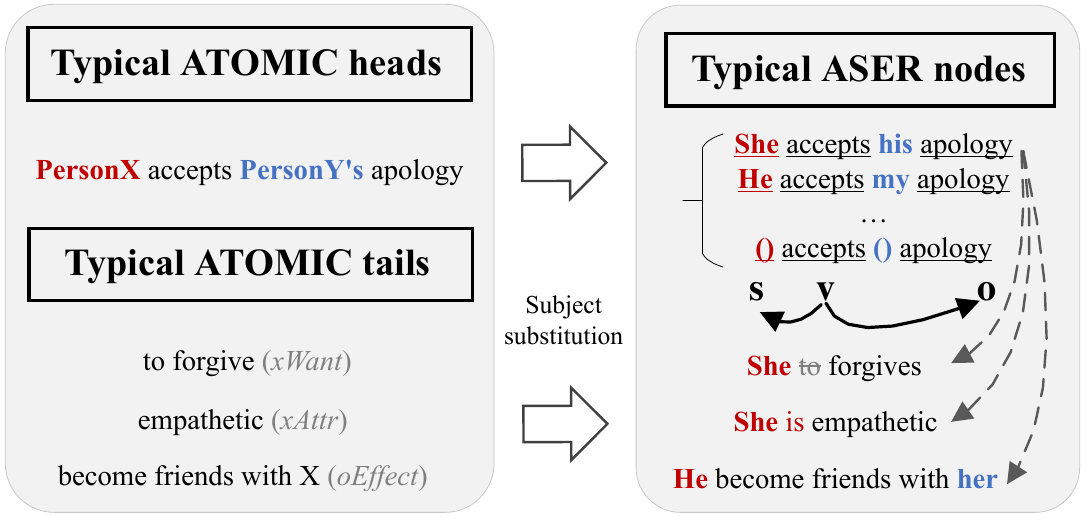}
    \caption{An illustration of the alignment between ATOMIC and ASER. We replace the placeholders ``\textit{PersonX}'' and ``\textit{PersonY}'' with concrete singular personal pronouns, and add subjects to ATOMIC tails to make them complete sentences. 
    } \label{fig:mapping-sketch}
    \vspace{-2ex}
\end{figure}

\subsection{Aligning ATOMIC and ASER}\label{section:mapping}

In ATOMIC, the nodes are eventualities with ``\textit{PersonX}'' and ``\textit{PersonY}'' as subjects or objects. 
However, in ASER, the corresponding eventualities are nodes with concrete personal pronouns, for example, \textit{I}, \textit{she}, \textit{Alex}, and \textit{Bob}.
In addition, as the tails in ATOMIC are written by human annotators, the formats can be arbitrary and sometimes subjects are missing in tails.
To effectively align the information in ATOMIC and ASER, based on the above observations, we propose best-effort rules to convert ATOMIC nodes into the format of ASER, as shown in Table~\ref{table:atomic-preprocess}. Examples of the mapping process are shown in Figure~\ref{fig:mapping-sketch}. After conducting the string substitution operations, we use the parser in ASER to parse the acquired text into standard ASER format. 

The mapping statistics are shown in Table \ref{table:mapping_stat}, where 
the average percentage of ATOMIC nodes that can be detected in ASER, denoted as coverage, is 62.9\%. 
It is worth noting that the relation with the highest coverage is \textit{xAttr}, where the tails are mostly adjectives. By adding a personal pronoun and a \textit{be} in front of the \textit{xAttr} tail, we can find most \textit{stative} eventualities in ASER.

\begin{figure*}[t]
    \centering
        \includegraphics[width=0.7\textwidth]{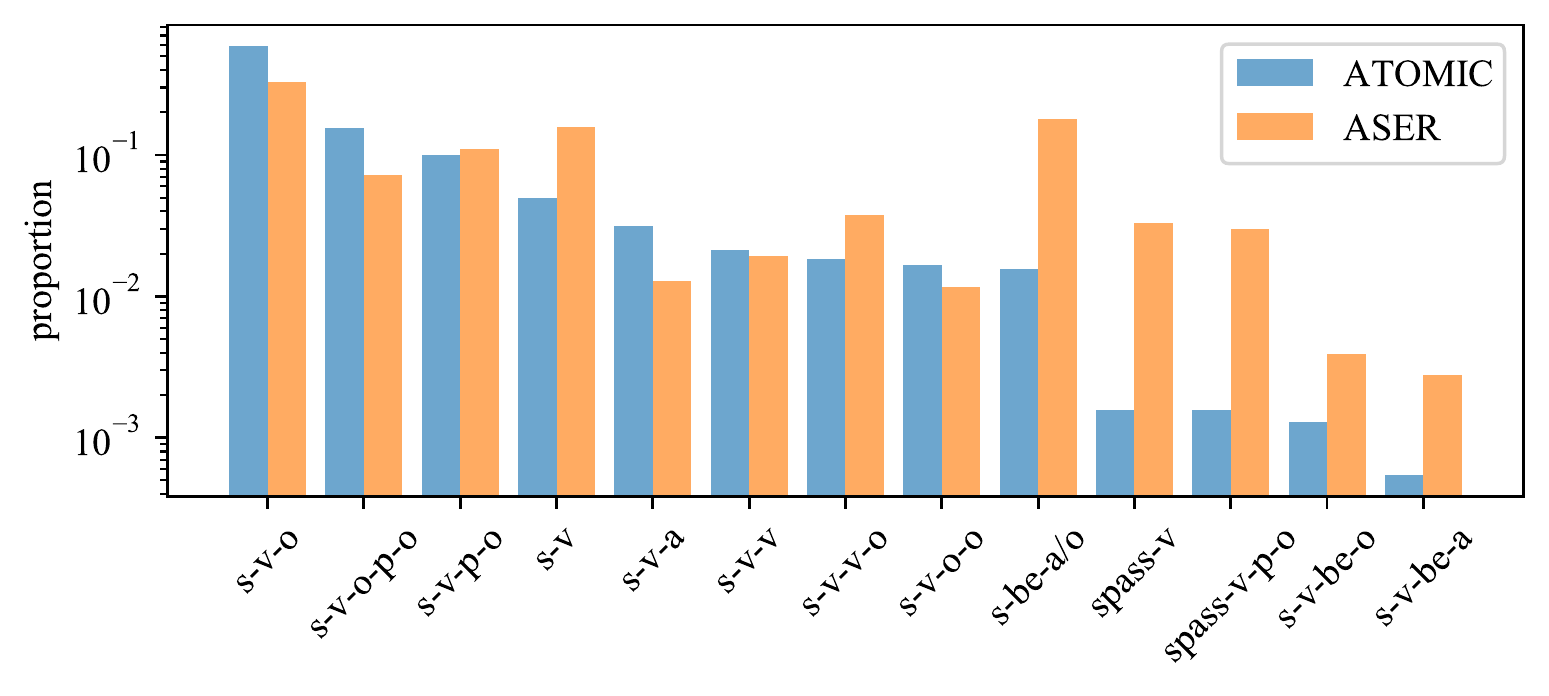}
    \vspace{-2ex}
    \caption{Pattern distribution of ATOMIC heads and eventualities in ASER.} \label{fig:pattern}
\end{figure*}

We further study the dependency pattern distribution of ATOMIC heads. The head events of ATOMIC are extracted from various corpora, including Google Ngrams and Wiktionary idioms. 
The definitions of events \cite{sap2019atomic} are similar with that in ASER.
We examine the coverage of their dependency patterns using the parser defined in ASER. There are 13 eventuality dependency patterns defined in ASER, as suggested in the paper \cite{zhang2020aser}, for example, s-v-o, s-v-o-p-o (`v' for normal verbs other than `be', `n' for nouns, `a' for adjectives, and `p' for prepositions.) The distribution of ATOMIC head patterns and ASER patterns is presented in Figure \ref{fig:pattern}. The Pearson $r$ between the distribution of ATOMIC pattern and ASER-core pattern is 0.8136, with $p<0.01$, showing consistency of ATOMIC and ASER. The syntactical patterns can be used to select eventualities when matching. For example, in ``\textit{xAttr}'' relation, we restrict the candidate tails in ASER to be of syntactical patterns ``s-v-a'' and ``s-v-o.''



\subsection{Discourse Knowledge Graph Preparation}\label{section:know-extract}

\begin{figure*}[htbp]
\centering
    \subfigure[\textit{Cause\_agent}]{
    \includegraphics[width=0.4\textwidth]{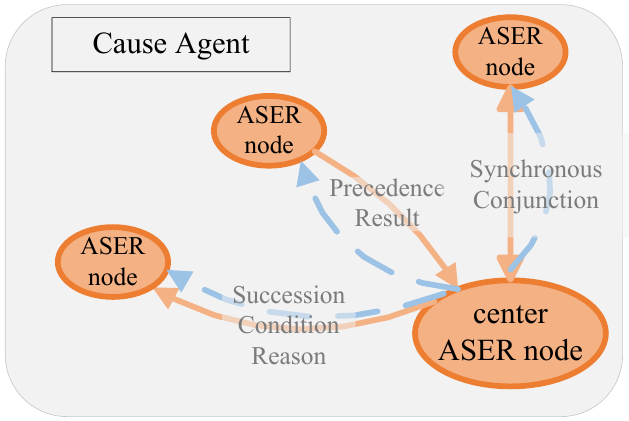}
    }
    \subfigure[\textit{Stative}]{
    \includegraphics[width=0.38\textwidth]{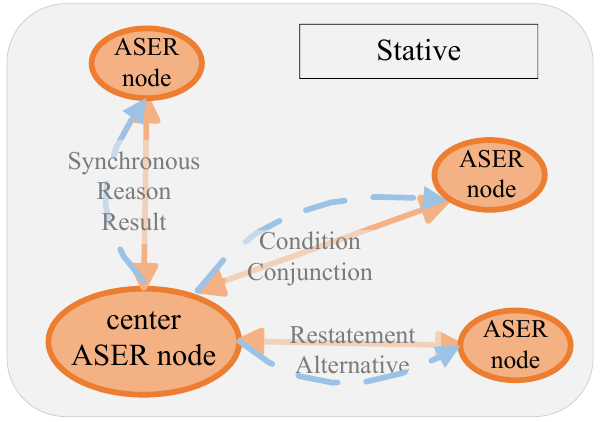}
    }
\centering
\subfigure[\textit{Effect\_theme}]{
\includegraphics[width=0.395\textwidth]{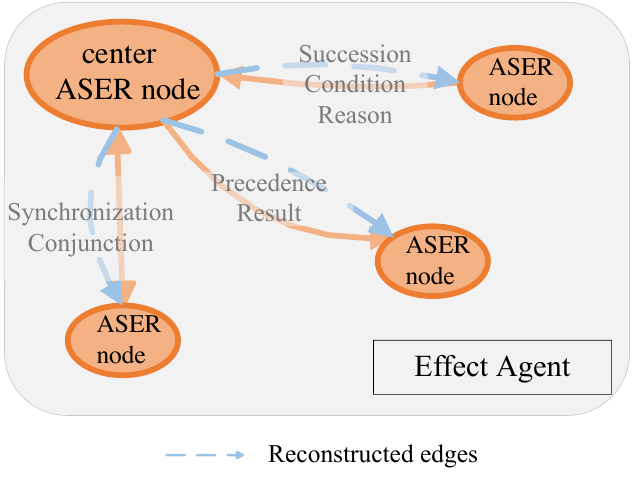}
}
\subfigure[\textit{Effect\_agent}]{
\includegraphics[width=0.39\textwidth]{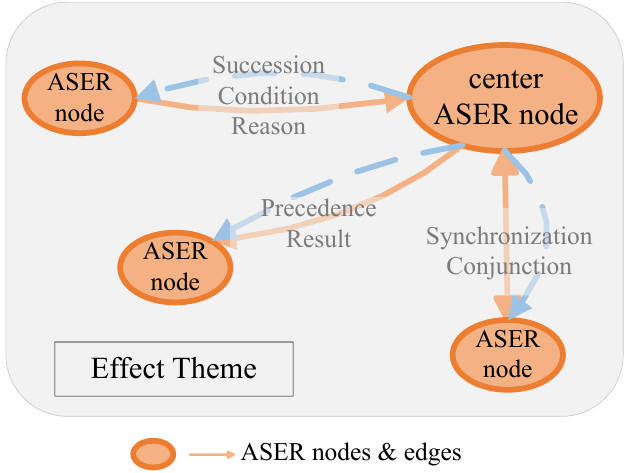}
}

\caption{Discourse knowledge extraction rules for different relation categories. The coral edges represent candidate ASER directed edges to be selected for a certain relation category. The dotted blue edges represent the reconstructed edges in $G_r$. }
\label{fig:aser-edge-rule}
\end{figure*}

We then introduce how to select candidate discourse edges from ASER. 
For a given node $u$ and a relation $r$, we find the edges based on the rule $D(u, r)$. As we are studying \textit{if-then} relations, the candidate discourse edges in ASER should be consistent with the order of time in the ATOMIC relation $r$. 
For example, for a commonsense tuple $(h, r, t)$ in the \textit{effect\_agent} category, the event $t$ is an effect of $h$ and thus $t$ should happen at the same time or after the event $h$. 
To retrieve ASER discourse edges with the same temporal logic, we first reconstruct an ASER subgraph by selecting specific edge types based on an ATOMIC relation $r$ with rules illustrated in Figure \ref{fig:aser-edge-rule}. 

\begin{table}[t]
\small
\begin{tabular}{m{0.35cm}<{\centering}|m{2.2cm}<{\centering} | m{4.5cm}}
  \toprule
  \multicolumn{2}{c|}{} & Mapping rules \\
  \hline
  \multicolumn{2}{c|}{Head} & Replace \textit{PersonX} and \textit{PersonY} with I/he/she/man/women/person \\
  \hline 
  \multirow{4}{0.7cm}{Tail} &\tabincell{c}{xWant/oWant/\\xIntent/xNeed} & Add a personal pronoun in front of the tail and remove the initial ``to''  \\
  &xEffect/oEffect&Add a personal pronoun in front of the tail \\
  &xReact/oReact&Add a personal pronoun and ``be'' in front of the tail \\
  &xAttr& Add a personal pronoun and ``be'' in front of the tail \\
  \bottomrule
 \end{tabular}
 \caption{Mapping rules from ATOMIC to ASER.}
 \label{table:atomic-preprocess}
  \vspace{-5ex}
\end{table}

We use the \textit{effect\_agent} category as an example. For a given node $u\in \mathcal{V}$, we select the directed  $(u, v)$ pairs from ASER, 
such that there exists either an edge $(u, v)\in \mathcal{E}$ where the edge types are among discourse relations \textit{Precedence} and \textit{Result},  
an edge $(v, u)\in \mathcal{E}$ where the edge types are among \textit{Succession}, \textit{Condition}, and \textit{Reason}, 
or there exists an $e \in \{(u, v), (v, u)\}$ such that the edge types of $e$ is among \textit{Synchronization} and \textit{Conjunction}. 
In this way, all the selected directed tuples $(u, v)$ represent the same temporal order as in the ATOMIC relation $r$.

In the next step, we need to distinguish the \textit{theme} categories from \textit{agent} categories. For relations under \textit{effect\_theme}, only eventuality pairs $(u, v)$ with different personal pronouns are selected as candidate knowledge, while for other \textit{agent}-based categories we select eventuality pairs with the same personal pronouns.
After this process, combined with all the mapped ATOMIC nodes, we collect all selected edges from $\mathcal{G}$,  to form an ASER-induced directed graph $G_r=(V_r, E_r)$ for each relation. $V_r$ is the set of vertices that includes both vertices from $\mathcal{G}$ and the aligned version of $\mathcal{C}$, and $E_r$ is the set of reconstructed edges according to the discourse knowledge selection rules defined above.
Here, an edge $(u,v)\in E_r$ can be viewed as a candidate ``\textit{if} $u$, \textit{then} $v$'' relation under $r$. 

After that, we aggregate the nodes in $G_r$ by conducting personal pronoun substitution. For the \textit{agent}-based relations, considering an edge $(u_r, v_r)\in E_r$, we replace the common personal pronouns in $u_r$ and $v_r$ as ``\textit{PersonX},'' to be consistent with the ATOMIC format. For other personal pronouns, we map them to ``\textit{PersonY}'' and ``\textit{PersonZ}'' according to the order of their occurrences. For the \textit{theme}-based relations, we replace the subject of $u_r$ with ``\textit{PersonX}'' and $v_r$ with ``\textit{PersonY}.'' After the personal pronoun substitution operation, we can acquire a unified discourse knowledge graph $G_r^c=(V_r^c, E_r^c)$ in the same format as ATOMIC. The corresponding statistics of all $G_r^c$ are shown in Table~\ref{table:mapping_stat}.

\begin{table}[t]
\renewcommand\arraystretch{1.0}
 \centering
 
 \small
 \begin{tabular}{l|c|cc|cc}
  \toprule
   & & \multicolumn{2}{c}{ATOMIC} & \multicolumn{2}{|c}{ASER $G_r^c$}\\
   \hline
   & Coverage(\%) & \#nodes &\#edges&\#nodes &\#edges  \\
  \hline
oEffect&31.1&25,328& 57,801&170,086&381,135\\
oReact&87.3&22,970& 59,839&95,169&320,543\\
oWant&61.6&38,892& 107,588&177,057&424,409\\
xAttr&95.8&32,959& 174,429&167,869&698,785\\
xEffect&33.1&43,840& 78,644&217,416&721,079\\
xIntent&33.8&33,789& 46,789&179,665&625,144\\
xNeed&52.9&51,206& 92,428&207,317&698,770\\
xReact&88.7&32,670& 99,162&145,216&528,918\\
xWant&58.8&61,149& 114,217&220,786&724,546\\

    \hline
    Head & 56.3 & - & - & -& -\\
    \hline
    Average & 62.9&38,089&92,322&175,620&569,259\\
    \bottomrule
  \end{tabular}
 \caption{Mapping statistics. The ATOMIC columns show the nodes and edges statistics of the graph produced by tuples in ATOMIC. The ASER $G_r^c$ column shows the statistics of the ASER-induced graph for a relation $r$ after personal pronoun aggregation. }\label{table:mapping_stat}
  \vspace{-5ex}
\end{table}

\subsection{Commonsense Knowledge Graph Population with \textsc{BertSAGE}}\label{section:BertSAGE}


In our framework, we train a CKGP model on the aligned graph $G_r^c$.
The basic goal of each step in CKGP is to classify whether a candidate discourse knowledge tuple $(u, v)\in E_r^c$ is a plausible \textit{if-then} commonsense knowledge under relation $r$. We use the commonsense tuples provided by ATOMIC as the seed ground truth edges.
For the negative examples, we explore several different sampling strategies: 
\begin{enumerate}[leftmargin=*]
\item\textbf{\textsc{Rand}} (\textsc{\textbf{Rand}om}): Randomly sample two nodes $(u, v)$ from $G_r^c$ such that $(u, v)\notin E_r^c$. 
\item \textbf{O} (\textbf{O}thers): Randomly sample two nodes $(u, v)$ from other relations such that $(u, v)\in E_{r'}^c, r'\in \mathcal{R},r'\neq r$. These negative samples will help the model to distinguish different commonsense relations. 
\item \textbf{I} (\textbf{I}nversion): Randomly sample a tuple $(u, v)\in E_r^c$ and add the inversion $(v, u)$ as negative samples. This is used to help the model understand the causal \textit{if-then} relationships, when the input tuples have similar semantic meanings  
\item \textbf{S} (\textbf{S}huffling ATOMIC): Randomly select $u$ from the set of ATOMIC heads, and $v$ from the set of ATOMIC tails under relation $r$. Add a negative sample if $(u, v)$ is not connected by an existing ATOMIC edge. This mechanism will prevent the model from assigning high scores only to nodes that have appeared in the ATOMIC training set. 
\end{enumerate}



To effectively encode both the semantic meaning of eventuality nodes and their neighbors on the overall graph, as shown in the right part of Figure~\ref{fig:model-overview}, we propose a model \textsc{BertSAGE} that contains two components:
(1) a node encoder based on BERT that embeds the semantic meaning of nodes; (2) a graph encoder that learns and aggregate relational information from the discourse graph.
The details are as follows.


\noindent $\bullet$ \textbf{Node encoder:}  We use the pre-trained language representation mode BERT \cite{devlin-etal-2019-bert} to encode all the nodes in the dataset. 
For a node $v=[w_1, w_2, \cdots, w_n]$ with $n$ word tokens, we add a [CLS] token in the beginning of each sentence as $w_0$ and a [SEP] token at the end of it as $w_{n+1}$. 
We denote the contextualized representation provided by BERT as $[\bm{e}_{w_0}, \bm{e}_{w_1}, \cdots, \bm{e}_{w_{n+1}}]$, $ \bm{e}_{w_i}\in \mathbb{R}^d$, where $d$ is the dimension of BERT embeddings, $\bm{e}_{w_0}$ and $\bm{e}_{w_{n+1}}$ are the embedding of [CLS] and [SEP] tokens, respectively. We then use the average pooling to acquire the final node representation as $\bm{e}_v=\sum_{i=0}^{n+1}\bm{e}_{w_i}/(n+2)$.

\noindent $\bullet$ \textbf{Graph encoder:} 
To effectively encode the semantics from neighbor events on the discourse graphs, we propose to use GraphSAGE \cite{hamilton2017inductive} to aggregate the neighbor information of a given node $v$. 

Given a node $v$, we first acquire its contextualized representation $\bm{e}_v$, and then calculate the embeddings of $v$'s neighbors in $G_r^c$, which are denoted as $\mathcal{N}(v)$. Here, $\mathcal{N}(v)$ is a fixed size neighbor set uniformly sampled from all the neighbors of $v$. The hidden representation after the GraphSAGE layer $\bm{h}_{v}$ is computed as follows:

\begin{align}
    \bm{h}_{\mathcal{N}(v)} &\leftarrow \small{\text{AGGREGATE}}(\{\bm{e}_u, \forall u\in \mathcal{N}(v)\})\label{eq:BertSAGE1}, \\
    \bm{h}_{v}&\leftarrow \sigma (\bm{W}\cdot \small{\text{CONCAT}}(\bm{h}_v, \bm{h}_{\mathcal{N}(v)})).
\end{align}

\noindent $\bullet$ \textbf{Output layer:} For an input candidate tuple $(u, v) \in G_r^c$, on top of the overall representation given by \textsc{BertSAGE} $[\bm{h}_{u}, \bm{h}_{v}]$, we apply an output layer $f_r(u, v) = \text{Softmax}([\bm{h}_{u}, \bm{h}_{v}] \bm{W}^{'\top}+\bm{b})$, $\bm{W}^{'}\in \mathbb{R}^{2\times d}, \bm{b}\in \mathbb{R}^2$ to make the final prediction.


\section{Experiments}\label{section:experiments}
We introduce the experimental settings and results of all the experiments in this section. Both learning and inference processes in Section~\ref{sec:task-definition} are studied here.
For the learning part, the task is a link prediction task and thus we evaluate the performance automatically using accuracy based on the existing annotated commonsense knowledge as positive examples and automatically sampled edges as negative examples.
For the inference part, as the goal to acquire novel commonsense knowledge is similar with previous works in ATOMIC~\cite{sap2019atomic} and COMET~\cite{bosselut-etal-2019-comet}, 
we adopt human evaluation to evaluate the quality of the newly acquired knowledge, and then use novelty and diversity as additional evaluation metrics accordingly. 

\begin{table*}[t]
  \centering
  \small
  \label{tab:LABEL}
  \begin{tabular}{l|ccccccccc}
  \toprule
    Model&oEffect&oReact&oWant&xAttr&xEffect&xIntent&xNeed&xReact&xWant \\
    \hline
    \textsc{Bert}&90.60&{97.05}&{93.95}&{96.21}& 87.85& 89.69& 89.93& {93.96} & 89.73\\
    \textsc{BertSAGE}&\textbf{91.10}*&\textbf{97.29}&\textbf{94.21}& \textbf{96.33}& \textbf{89.49}*& \textbf{90.48}*& \textbf{91.10}*& \textbf{94.02}& \textbf{90.91}*\\
    \bottomrule
  \end{tabular}
  \caption{Evaluations on the CKGP
  link prediction experiments. We report the accuracy in test set here as the number of positive and negative samples are balanced.
  * after bold figures indicates that the improvement of \textsc{BertSAGE} model is significant with z-test $p<0.05$. 
  }\label{table:ckg-pop}
\end{table*}

\subsection{Learning for CKGP}
\subsubsection{Settings}

We first train the \textsc{BertSAGE} model for Commonsense Knowledge Graph Population. 
We evaluate the performance of link prediction using accuracy.
We use the edges derived from ATOMIC as positive examples. 20\% of the negative examples are randomly sampled using \textbf{O}, 10\% of them using \textbf{I}, and the rest using \textbf{\textsc{Rand}}, as defined in Section~\ref{section:BertSAGE}. Detailed ablation studies about negative sampling techniques are presented in Section~\ref{section:ablation}.
Considering that $G_r^c$ is much larger than ATOMIC, we restrict the size of $G_r^c$ in the following ways. (1) We only select the subgraph of $G_r$ which is induced by the one-hop neighbors of all the ATOMIC nodes. (2) For the subgraph acquired in the first step, we add two hop neighbors into the graph for the nodes whose degrees are less than a threshold $k$. $k$ is set to 20 in practice.
We use \textit{bert-base-uncased}\footnote{\url{https://github.com/huggingface/transformers}}~\cite{devlin-etal-2019-bert} as the encoding layer for the classification model and the dimension of the hidden embeddings is 768.
For the neighbor function $\mathcal{N}(u)$, we set the neighbor size to be 4. The batch size is set to be 64. We use the train/dev/test split defined in ATOMIC \cite{sap2019atomic}. 
To clearly show the contribution of the proposed \textsc{BertSAGE} model, we compare it with a modified version of KG-Bert \cite{yao2019kg}, denoted as \textsc{Bert} baseline for short, on the link prediction task. The only difference between \textsc{BertSAGE} and \textsc{Bert} is that we incorporate the semantic about neighboring events to get the final representation in \textsc{BertSAGE}.
We use the same setting for \textsc{Bert} as defined above except for the graph module.

\subsubsection{Result}


We report the accuracy of the CKGP models on the test set with prepared negative edges.
From the results in Table \ref{table:ckg-pop}, we can see that adding a GraphSAGE layer over the \textsc{Bert} baseline will improve the classification results on all relation types. 
These results prove our assumptions that adding information about the neighbor events on the discourse graph can help generate better event representations.
Among nine relations, the improvements are significant with z-test $p<0.05$ on five types.
One interesting finding is that this improvement is in positive correlation with the graph complexity in Table \ref{table:mapping_stat}. In general, GraphSAGE will contribute more to the performance when the graph is more complex.

\subsection{Inference for CKGP}\label{sec:experiments-ckgc}

\subsubsection{Settings}

We evaluate the capability of the above \textsc{BertSAGE} model with the inference part for acquiring new commonsense knowledge in CKGP. 
The goal of the inference part is similar with that in COMET. 
As there is no ground truth for the newly generated nodes or edges, we conduct human evaluation for the quality. 
Besides accuracy, we also use automatic metrics related to novelty and diversity to demonstrate the properties of the acquired commonsense knowledge.
While COMET uses a neural generation method to generate tails, in DISCOS we use \textsc{BertSAGE} to rank the candidates provided in ASER given heads and relations.
Similar with COMET, for the \textit{existing head} setting introduced in Section~\ref{sec:task-definition}, 
we propose to evaluate the acquired commonsense knowledge from all three perspectives, i.e., quality, novelty, and diversity. While for the \textit{novel head} setting, as the heads are already novel, we only evaluate the Quality of the retrieved knowledge.

\begin{table*}[t]
  \centering
  \small
  \label{tab:LABEL}
  \begin{tabular}{l|cccccccccc}
  \toprule
    Model&oEffect&oReact&oWant&xAttr&xEffect&xIntent&xNeed&xReact&xWant\\
    \hline
    COMET* (beam-search top 10) & 29.0& 37.7& 44.5& 57.5& 55.5& 68.3& 64.2& 76.2& 75.2 \\
    COMET (beam-search top 10)&59.8&\textbf{69.6}&69.0&\textbf{77.7}*&\textbf{75.4}*&86.2&80.7&\textbf{75.6}*&\textbf{78.9}*\\
    DISCOS (top 10) &\textbf{68.3}*&67.1&\textbf{69.9}&66.7&60.9&\textbf{87.8}&\textbf{84.9}&68.4&73.4\\
    \hline
    Human* & 84.6& 86.1& 83.1& 78.4& 83.9& 91.4& 82.0& 95.2& 90.0 \\ 
    
    
    \bottomrule
  \end{tabular}
  \caption{Human evaluation on quality of the inference process in CKGP, under the \textit{existing head}
  setting (given $(h, r)$ to predict $t$.) COMET* represents the results provided by the original paper of COMET. In the last row, we report the human evaluation results of the gold ATOMIC in the COMET paper \cite{bosselut-etal-2019-comet}. * after bold figures indicates the scores are significantly superior to the other measured by z-test with $p<0.05$. }\label{table:human-eval}
\end{table*}

\begin{table*}[t]
  \centering
  \small
  \label{tab:LABEL}
  \begin{tabular}{l|p{0.46cm}<{\centering}p{0.46cm}<{\centering}|p{0.46cm}<{\centering}p{0.46cm}<{\centering}|p{0.46cm}<{\centering}p{0.46cm}<{\centering}|p{0.46cm}<{\centering}p{0.46cm}<{\centering}|p{0.46cm}<{\centering}p{0.46cm}<{\centering}|p{0.46cm}<{\centering}p{0.46cm}<{\centering}|p{0.46cm}<{\centering}p{0.46cm}<{\centering}|p{0.46cm}<{\centering}p{0.46cm}<{\centering}|p{0.46cm}<{\centering}p{0.46cm}<{\centering}}
  \toprule
  
\multirow{2}{2cm}{Model} & \multicolumn{2}{c}{oEffect} &\multicolumn{2}{c}{oReact} &\multicolumn{2}{c}{oWant} &\multicolumn{2}{c}{xAttr} &\multicolumn{2}{c}{xEffect} &\multicolumn{2}{c}{xIntent} &\multicolumn{2}{c}{xNeed} &\multicolumn{2}{c}{xReact} &\multicolumn{2}{c}{xWant}\\\cline{2-19}
&$NT_o$&$NU_o$&$NT_o$&$NU_o$&$NT_o$&$NU_o$&$NT_o$&$NU_o$&$NT_o$&$NU_o$&$NT_o$&$NU_o$&$NT_o$&$NU_o$&$NT_o$&$NU_o$&$NT_o$&$NU_o$\\ 
\hline
COMET @ 1 & 0.0 &  0.0&  0.0 &  0.0&  2.2 &  10.4&  0.2 &  0.7&  2.0 &  8.3&  4.1 &  9.9&  12.0 &  29.3&  0.0 &  0.0&  9.4 &  18.1 \\
\name{} @ 1 &\textbf{61.2} & \textbf{65.0}& \textbf{15.5} & \textbf{36.3}& \textbf{43.2} & \textbf{56.7}& \textbf{6.8} & \textbf{17.2}& \textbf{45.3} & \textbf{54.2}& \textbf{38.5} & \textbf{61.6}& \textbf{29.7} & \textbf{44.0}& \textbf{5.5} & \textbf{21.4}& \textbf{25.6} & \textbf{45.8} \\
\hline
COMET @ 2 & 7.5 &  23.9&  0.0 &  0.0&  3.5 &  17.8&  0.1 &  0.4&  2.3 &  9.3&  6.2 &  14.8&  11.8 &  32.0&  0.0 &  0.0&  10.2 &  20.0 \\
\name{} @ 2 &\textbf{58.4} & \textbf{65.9}& \textbf{13.7} & \textbf{33.8}& \textbf{45.1} & \textbf{63.0}& \textbf{7.1} & \textbf{20.1}& \textbf{45.3} & \textbf{58.6}& \textbf{39.5} & \textbf{63.0}& \textbf{30.5} & \textbf{49.3}& \textbf{6.2} & \textbf{26.5}& \textbf{26.3} & \textbf{49.1} \\
\hline
COMET @ 5 & 12.9 &  30.3&  0.1 &  1.0&  7.5 &  25.4&  0.1 &  0.7&  5.3 &  17.0&  8.6 &  21.4&  15.8 &  35.5&  0.1 &  0.9&  11.5 &  26.5 \\
\name{} @ 5 &\textbf{59.9} & \textbf{72.8}& \textbf{16.5} & \textbf{38.8}& \textbf{49.9} & \textbf{69.6}& \textbf{8.9} & \textbf{25.8}& \textbf{50.2} & \textbf{65.3}& \textbf{44.9} & \textbf{68.8}& \textbf{38.2} & \textbf{59.5}& \textbf{6.9} & \textbf{33.7}& \textbf{32.2} & \textbf{55.1} \\
\hline
COMET @ 10 & 16.8 &  40.4&  0.4 &  4.9&  9.8 &  32.4&  0.1 &  0.8&  8.0 &  24.1&  12.7 &  31.2&  18.6 &  41.0&  0.4 &  4.7&  12.3 &  30.5 \\
\name{} @ 10 &\textbf{62.9} & \textbf{76.2}& \textbf{22.5} & \textbf{50.4}& \textbf{55.8} & \textbf{75.4}& \textbf{12.0} & \textbf{30.4}& \textbf{54.5} & \textbf{71.1}& \textbf{51.7} & \textbf{74.1}& \textbf{44.2} & \textbf{66.2}& \textbf{9.1} & \textbf{42.8}& \textbf{38.1} & \textbf{62.0} \\
    \bottomrule
  \end{tabular}
  \caption{Novelty on the test set grouped by different relations, under \textit{existing head} setting of the inference process. @ $k$ means we evaluate the top $k$ generation or retrieval for a given head $h$. All improvements by \name{} are significant with z-test $p$<0.05. }\label{table:novelty}
\end{table*}

\begin{table}[t]
  \centering
  \small
  \label{tab:LABEL}
  \begin{tabular}{l|cc|cc}
  \toprule
    & \multicolumn{2}{c|}{dist-1} & \multicolumn{2}{c}{dist-2}\\
  \hline
   relations & COMET & DISCOS & COMET & DISCOS \\
   
    \hline
oEffect& 60.3 & \textbf{66.7} & 76.3 & \textbf{89.3} \\
oReact& \textbf{35.5} & 33.5 & 13.5 & \textbf{35.9} \\
oWant& 46.6 & \textbf{69.0} & 84.1 & \textbf{93.8} \\
xAttr& 8.3 & \textbf{26.0} & 4.2 & \textbf{27.4} \\
xEffect& 58.4 & \textbf{67.2} & 81.8 & \textbf{90.4} \\
xIntent& 42.9 & \textbf{61.5} & 75.7 & \textbf{87.3} \\
xNeed& 41.4 & \textbf{63.6} & 75.7 & \textbf{88.4} \\
xReact& 27.1 & \textbf{29.3} & 12.1 & \textbf{32.9} \\
xWant& 42.2 & \textbf{65.3} & 78.7 & \textbf{91.5} \\
\hline
Average& 38.3 & \textbf{52.9} & 55.0 & \textbf{70.0} \\
    \bottomrule
  \end{tabular}
  \caption{Diversity grouped by all the relations for the \textit{existing head} setting in the inference process. We report  the  diversity of  top 10 generaion or retrieval of  COMET and DISCOS. Dist-$k$ indicates the proportion of unique $k$-grams.}\label{table:diversity}
  \vspace{-3ex}
\end{table}

\begin{enumerate}[leftmargin=*]
 \item \textbf{Quality}:  We evaluate the quality of acquired commonsense knowledge using annotators from  Amazon Mechanical Turk (AMT.)
    For each relation in ATOMIC, we randomly sample 50 head events from the testing set and ask the annotators if they think the generated tuple makes sense.
    For COMET, we use beam 10 top 10 as the decoding mechanism to generate 10 commonsense knowledge for each head event. For DISCOS, we select the tuples ranked top 10 by the \textsc{BertSAGE} model.
    \item \textbf{Novelty}: We first evaluate the novelty of acquired commonsense knowledge with two novelty indicators, the proportion of generated tails that are novel ($NT_t$), and the proportion of novel tails in the set of all the unique generated tails ($NU_t$.)
    \item \textbf{Diversity}: Last but not least, considering that the novelty is evaluated based on string match, which cannot effectively distinguish whether a system is generating many different novel concepts or just similar but not identical concepts.
    Following previous works \cite{li2015diversity, zhang2018generating}, we report diversity indicator dist-1 and dist-2, the proportion of distinct unigrams and bigrams among the total number of generated unigrams and bigrams. We evaluate the diversity of generated knowledge given the same head and relation and calculate the average among all the heads.
\end{enumerate}

For COMET, we use the public available official implementation\footnote{\url{https://github.com/atcbosselut/comet-commonsense}}. All the experimental settings are the same as in the original paper. Similar with the decoding mechanisms in the COMET paper, we use beam search top $k$ to retrieve $k$ generated tails.

\subsubsection{Result}

The overall quality\footnote{We present the original human annotation results from the ATOMIC paper as a reference. However, as we employ different annotators, they are not comparable with our results. }, novelty, and diversity of COMET and DISCOS are shown in Table \ref{table:human-eval}, \ref{table:novelty}, and \ref{table:diversity}, respectively. 
From the results, we can make the following observations.
Based on our crowd-sourcing results, DISCOS can achieve comparable or better quality on \textit{effect\_theme} relations (\textit{oEffect}, \textit{oReact}, and \textit{oWant}) and \textit{cause\_agent} relations (\textit{xIntent} and \textit{xNeed}) among the nine relations. 
The results indicate that rich commonsense knowledge is indeed covered by the discourse graph and the proposed DISCOS framework can effectively discover them.
At the same time, we also notice that DISCOS can significantly outperform COMET in terms of the novelty.
For example, for some relations like \textit{xAttr}, \textit{oReact}, and \textit{xReact}, COMET hardly generate novel tails despite increasing the size of beam search while a large portion of the DISCOS knowledge is novel.
One reason behind is that COMET fits the training data too well and the training set is similar to the test set. As a result, it tends to predict the concepts it has seen in the training set rather than something new.
Last but not least, similar to the novelty, DISCOS also outperforms COMET in terms of the diversity, which is mainly due to the limitation of beam search as it often generates very similar sentences.
As DISCOS is a classification model rather than a generation model, it does not suffer from that problem.
To conclude, compared with COMET, DISCOS can acquire much more novel and diverse commonsense knowledge with comparable quality.



\begin{table*}[t]
\centering
\small
\begin{tabular}{l|cccccccccc}
\toprule
 & oEffect & oReact & oWant & xAttr & xEffect & xIntent & xNeed & xReact & xWant\\
\hline
DISCOS (novel heads and tails) & 70.2 & 63.2 & 59.4 & 69.2 & 78.2 & 65.8& 67.8& 80.0& 49.2\\ 
\bottomrule
\end{tabular}
\caption{Human annotation on quality for the \textit{novel head} setting (given $r$ to retrieve plausible $(h, t)$ pairs in DISCOS.)}\label{table:human-eval-kbp}
\end{table*}

To further demonstrate that DISCOS has the potential to acquire the commonsense knowledge without the help of human defined heads, we evaluate it with the \textit{novel head} setting.
Here, only the relation $r$ is provided and the model is asked to retrieve the novel $(h, t)$ pairs from ASER.
Specifically, we select the tuples scored higher than 0.5 by the \textsc{BestSAGE} model, and randomly sample 100 tuples form each relation for human evaluation.
To make sure the acquired knowledge is not observed by the model, only novel concepts are evaluated.

From the results in Table \ref{table:human-eval-kbp}, we can see the potential of DISCOS in directly mining high-quality novel commonsense knowledge from the raw graph of ASER.
For example, it achieves over 70\% accuracy on three relations ( \textit{``oEffect''}, \textit{``xEffect''}, and \textit{``xReact''}.)
Following this experimental setting, we successfully convert ASER into a large scale commonsense knowledge graph DISCOS-ATOMIC, which contains 3.4 million complex commonsense knowledge in the format of ATOMIC, without using any additional annotation effort. 

\section{Ablation Study}\label{section:ablation}
In this section, we will present ablation studies on the effects of different negative sampling strategies on the link prediction part of the CKGP task.
We tried to use aforementioned combinations in Section~\ref{section:BertSAGE} to generate the negative examples for both the training and testing set, 
and present the results of link prediction accuracy\footnote{We select the \textit{xWant} relation as an example.} 
on the test set in Table \ref{table:ablations}.
Specifically, we tried the following combinations:
\begin{enumerate}[leftmargin=*]
    \item \textsc{Rand}: All the negative examples are sampled randomly from the whole graph.
    \item O20: 20\% of the negative examples are sampled using the mechanism \textbf{O}.
    \item O20+I10: 20\% of the negative examples are sampled using the mechanism \textbf{O} and 10\% from the mechanism \textbf{I}.
    \item O20+I10+S10: 20\% of the negative examples are sampled using the mechanism \textbf{O}, 10\% from the mechanism \textbf{I}, and 10\% from the mechanism \textbf{S}. 
\end{enumerate}
We highlight the accuracy ranked highest on O20+I10+S10 test set, the hardest negative example set. 
From the result we can see that, even though the \textsc{Rand} achieves comparable performance on the simple test set \textsc{Rand}, it suffers a significant performance drop on the other harder ones.
The reason behind is that the randomly selected negative examples can only help the model to distinguish the ATOMIC positive examples rather than distinguish the commonsense.
This ablation study also demonstrates the importance of including more diverse negative example generation strategies to cover more signals we want the model to learn.
In the end, we choose to use O20+I10 negative sampling for training in our final model.
\begin{table}[t]
\renewcommand\arraystretch{1.0}
 \centering
 \small
 \begin{tabular}{p{1.8cm}<{\centering}| p{0.9cm}<{\centering} p{1.1cm}<{\centering} p{0.9cm}<{\centering} p{1.1cm}<{\centering} }
  \toprule
 \diagbox{\small{Test}}{\small{Train}}&  \textsc{Rand} &  O20&  O20+I10 &\tabincell{c}{O20+I10\\+S10} \\
  \hline
    \textsc{Rand} &94.40&93.65&93.50&90.88\\
   \textsc{O20} &87.46&91.16&90.93&89.12 \\
   \textsc{O20+I10} &87.16&90.72&90.92&89.17\\
   O20+I10+S10&82.80&86.49&\textbf{86.85}&86.53\\
  \bottomrule
 \end{tabular}
 \caption{Ablation study on different negative sampling methods under \textit{xWant} relation, trained using \textsc{BertSAGE} model. We report the accuracy of testing set here using the link prediction task in CKGP. }\label{table:ablations}
\vspace{-5ex}
\end{table}

\section{Case Study and Discussion}\label{section:case-discussion}

In this section, we present some case studies of the generated commonsense knowledge and have discussions about our approach.

\subsection{Case Study}

Some examples of the generation results of COMET and the retrievals from DISCOS are presented in Table \ref{table:case-study} and \ref{table:case-study-novel-head}, where the former one is under the setting of COMET and the latter one is under the \textit{novel head} setting where commonsense tuples with novel heads are produced. From Table \ref{table:case-study}, we can see that sometimes the generations of COMET are very similar for the same head event. The difference between multiple generations is only the singular or plural form of verbs.  This is mainly due to the drawback of \textit{selection bias} in machine learning, making neural generation models harder to generate novel outputs. For the retrieved results by DISCOS, the diversity and novelty can be inherently improved, while staying comparable accuracy.
In Table~\ref{table:case-study-novel-head}, ``X watches the World Cup'' does not appear in the head part of the ATOMIC dataset, and DISCOS can provide plausible \textit{xAttr} tail of it, which is ``X is excited.''

\subsection{Effects of ASER}

As stated in the quality analysis in Section \ref{sec:experiments-ckgc}, the performance of DISCOS on \textit{effect\_theme} categories (\textit{oEffect}, \textit{oReact}, and \textit{oWant}) is better or comparable with the COMET baseline.
This is because there are relatively fewer annotations for the \textit{effect\_theme} relations in ATOMIC. 
For \textit{oEffect}, \textit{oReact}, and \textit{oWant} relations, the average number of ATOMIC tails per head are  1.5, 1.4, 2.2, respectively, compared with 4.2 tails per head for other \textit{agent} (X)-driven relations. 
The performance of COMET drops on the \textit{theme} (Other)-driven relations as there are fewer training data, which is consistent with the findings in COMET paper \cite{bosselut-etal-2019-comet}.
DISCOS can fill the gap of limited training data by finding explicit candidate discourse knowledge from ASER, thus resulting in better or comparable performance. 
Next, the performances on relations in \textit{cause\_agent} category are also improved compared to COMET. These relations require the tails to happen ahead of the base event. But under the \textit{if-then} knowledge framework of COMET, a tail is generated after feeding the head and relation into the unidirectional language model, which is opposite from the definition of \textit{cause\_agent} in chronological order.
From the case studies of \textit{xIntent} and \textit{xNeed} relations in Table \ref{table:case-study}, we could also see that the COMET model sometimes confuses the causes and effects, i.e., the generated tails are not the \textit{causes} of the head  events as they are supposed to be, but the \textit{effects}. For example, for \textit{xNeed} relation in Table \ref{table:case-study}, \textit{if} ``X want to sleep'' \textit{then} ``X go to bed'' is a plausible sentence, but is not a plausible \textit{xNeed} commonsense knowledge, as ``X go to bed'' is the \textit{effect} of the base event instead of the \textit{cause}.
DISCOS can handle this situation well by selecting ASER edges with exactly the same chronological order as the definition of each relation. 
Moreover, Table~\ref{table:case-study-novel-head} demonstrates that by incorporating ASER, we can overcome the drawbacks of previous generation based method that they can only conduct reasoning on pre-defined heads. In ASER, every discourse edge can be viewed as a potential commonsense relation. Using a discriminative model like \textsc{BertSAGE}, we can have a more scalable method for acquiring novel commonsense knowledge.

\begin{table}[t]
    \centering
    \includegraphics[width=0.5\textwidth]{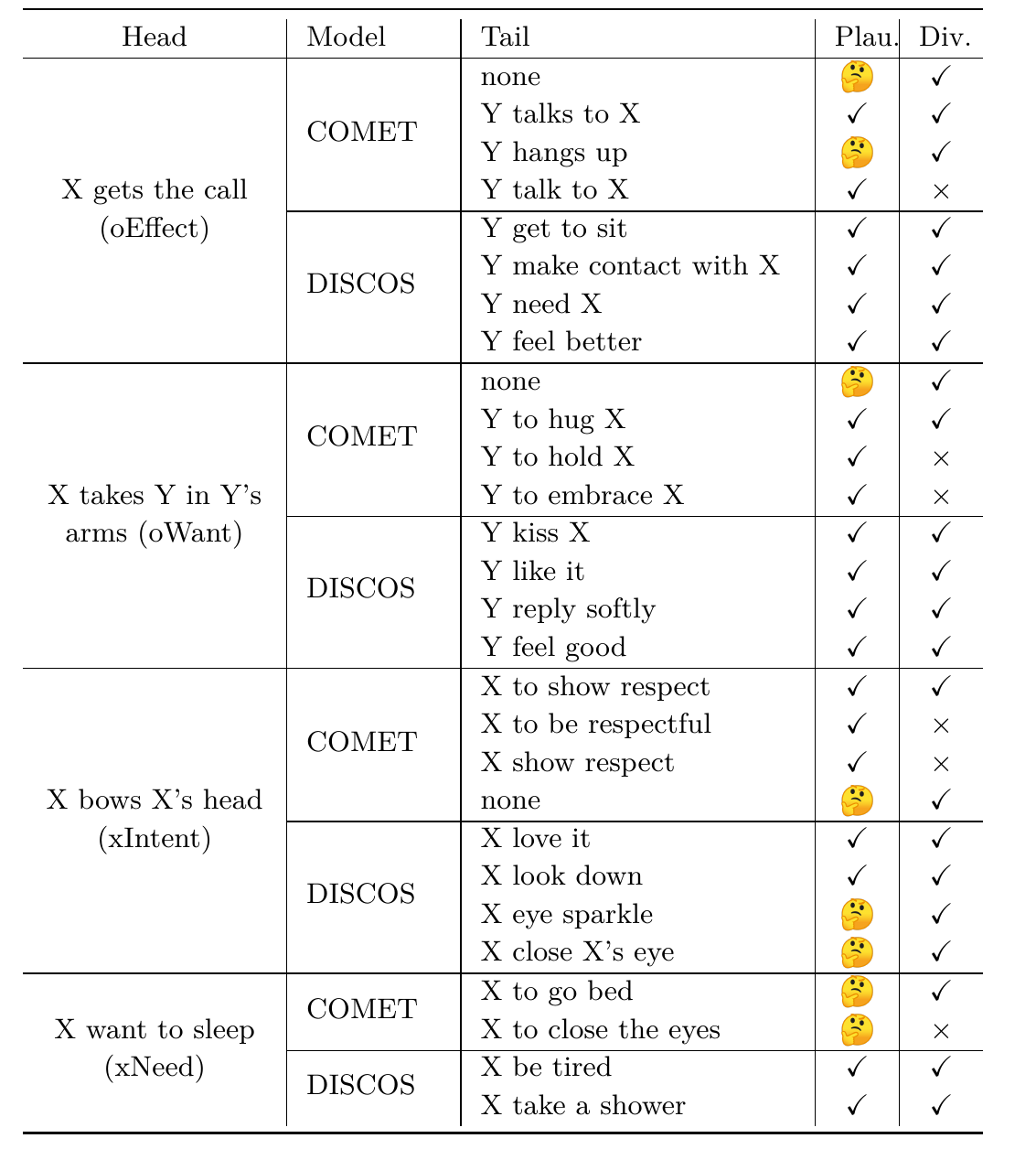}
    \caption{Case study of the \textit{existing head} setting of CKGP inference. Plau. indicates the plausibility of the tuple, and Div. indicates whether the generated tail is a diverse generation. For example, the tail ``Y talk to X'' in the \textit{oEffect} relation would not be considered as a diverse generation, as there exists a similar ``Y talks to X'' ahead of it. } \label{table:case-study}
    \vspace{-5ex}
\end{table}

\begin{table}[]
\centering
\small
\begin{tabular}{l|c|l}
\toprule
Head & Relation & Tail \\
\hline
\rowcolor{Gray}
X get pregnant & oEffect & Y have to watch \\
\rowcolor{Gray}
X sit down & oEffect & Y move a few pace away \\
X be very professional & oReact & Y be very satisfied \\
X visit Y & oReact & Y be quite happy \\
\rowcolor{Gray}
X give advice & oWant & Y follow it \\
\rowcolor{Gray}
X strike Y & oWant & Y seize X \\
X should rest & xAttr & X be tired \\
X watch the world cup & xAttr & X be excited \\
\rowcolor{Gray}
X have X's son & xEffect & X get pregnant again \\
\rowcolor{Gray}
X hurt Y feeling & xEffect & X again apologize \\
X take the gun & xIntent & X kiil Y \\
X have to go & xIntent & X realize something \\
\rowcolor{Gray}
X be in danger & xNeed & X need someone \\
\rowcolor{Gray}
X stand still & xNeed & X hold in hand \\
X keep worry & xReact & X be concerned \\
X look back & xReact & X be proud of career \\
\rowcolor{Gray}
X work in mall & xWant & X get a discount \\
\rowcolor{Gray}
X be hungry & xWant & X grab a bite\\
\bottomrule
\end{tabular}%
\caption{Case study of the \textit{novel head} setting of \name{} where commonsense tuples with novel heads are produced. Two generated samples from \name{} for each relation are presented.  }
\label{table:case-study-novel-head}
\end{table}

\section{Conclusion} \label{section:conclusion}

In this paper, we propose \name{}, a novel framework that populates the inferential commonsense knowledge in ATOMIC to an eventuality-centric discourse knowledge graph ASER.
The \textsc{BertSAGE} model can be served as a general method for such knowledge population task over graph.
Experimental results have shown that we can retrieve much more novel and diverse \textit{if-then} commonsense knowledge from ASER with high quality comparable with neural text generation models. This approach shows promising future for converting cheap discourse knowledge into expensive commonsense knowledge.

\section*{ACKNOWLEDGMENTS}

This paper was supported by the NSFC Grant (No. U20B2053) from China, the Early Career Scheme (ECS, No. 26206717),  the General Research Fund (GRF, No. 16211520), and the Research Impact Fund (RIF, No. R6020-19) from the Research Grants Council (RGC) of Hong Kong, with special thanks to the Huawei Noah’s Ark Lab for their gift fund.


\bibliographystyle{ACM-Reference-Format}
\bibliography{Reference}

\end{document}